# Age prediction using a large chest X-ray dataset

Alexandros Karargyris, Satyananda Kashyap, Joy T Wu, Arjun Sharma, Mehdi Moradi, Tanveer Syeda-Mahmood
IBM Research, Almaden, San Jose, CA 95120

## ABSTRACT

Age prediction based on appearances of different anatomies in medical images has been clinically explored for many decades. In this paper, we used deep learning to predict a person's age on Chest X-Rays. Specifically, we trained a CNN in regression fashion on a large publicly available dataset. Moreover, for interpretability, we explored activation maps to identify which areas of a CXR image are important for the machine (i.e. CNN) to predict a patient's age, offering insight. Overall, amongst correctly predicted CXRs, we see areas near the clavicles, shoulders, spine and mediastinum being most activated for age prediction, as one would expect biologically. Amongst incorrectly predicted CXRs, we have qualitatively identified disease patterns that could possibly make the anatomies appear older or younger than expected. Further technical and clinical evaluation would improve this work. As CXR is the most commonly requested imaging exam, a potential use case for estimating age may be found in the preventative counselling of patient health status compared to their age-expected average, particularly when there is a large discrepancy between predicted age and the real patient age.

## 1. INTRODUCTION

Using medical images to estimate a person's or an organ's age is not new and often can be useful for clinical and forensic purposes. For example, since 1937, there has been work using hand X-rays to estimate a person's bone age to evaluate endocrine growth disorders in the pediatric population[1]. Other analogous examples include T-scores for bone density with DEXA scans (decrease with age) and calcium scores for coronary arteries in computer tomography (CT) scans (increase with age). Not infrequently, radiologists may also report that a patient's brain CT has "chronic ischemic microvascular changes and atrophy out of proportion to the patient's age". In other words, various medical imaging modalities often contain visual features about a person's internal anatomical structures or organs. These features are apparent to the human eyes and often have some correlation pattern with the expected biological age.

This observed correlation between imaging visual features and a person's age makes the problem potentially solvable and interesting to computer vision researchers. We imagine that as computer vision research moves towards analyzing multiple imaging modalities (e.g. X-rays, CT, MR, etc.) of ever-increasing image quality, one potential useful output would be the computer's estimation of the patient's age at the person level and potentially for all the different organs individually.

Recent promising work in the area includes the automatic pediatric bone age assessment by computers, a machine learning challenge organized by the Radiological Society of North America (RSNA)[2]. Hand X-rays are particularly useful in estimating bone age in the pediatric population, where there is a specific expected sequence of ossification events that occur in different locations in the hands and wrists at different ages before the age of 18. The bone age estimation information has been used either clinically to guide diagnosis and treatment for children with growth disorders, or forensically in the court to establish whether a refugee or undocumented person is a legal minor or not[3, 4]. Varying computer vision work has also been done in other imaging modalities, such as automatic calcium scoring prediction. These organ-specific age estimations information could be used clinically to counsel patients or implement preventative measures aimed at reducing the risk of age-related comorbidities. For example, if a chest CT shows increased calcium scores or a radiologist observed "calcific atherosclerosis out of proportion to the patient's age", the patient would be counseled to avoid typical causes of aging, such as smoking, obesity and alcohol consumption. Similarly, a low bone density T-score might prompt a doctor to start the patient on fracture preventative management, such as exercise, hormone replacement or starting on a bisphosphate.

Chest X-rays (CXR) are the most commonly requested radiology modality. However, there is very limited work in the literatures showing age estimation using this modality in both the clinical and computer vision realm. We identified a single clinical publication by Gross BH et. al. [5] in which 4 radiologists were asked to estimate the age from CXRs. Results showed statistically significant differences among them. Interestingly, observer experience did not correlate with accuracy of patient age estimation. In another related work using Chest CT, Hochhegger et. al.[6] described how age affects radiograph findings and how to differentiate between the normal aging process and the onset of diseases. The scarcity of work is likely because, as a cheaper and quick imaging exam, CXRs are more often used for triaging and screening than final diagnostic purposes. They capture the image of the whole thorax but do not give the highest imaging detail about any one organ compared to other modalities like CT and MR. In [7] Barrès *et. al.*, presented clinical work in 55 chest radiographs that were obtained during routine autopsies. Age estimation was approximated through manual evaluation and grading (1-5) of five features: bone demineralization, fusion of the pieces in the manubrium, rib-to- cartilage attachment charges, cartilage mineralization, and cartilage-to-sternum attachment.

Nevertheless, visually, we still expect imaging features that would correlate with broader age groups in CXRs. For example, the clavicles and spine are visualized on CXRs and have some age-specific appearances, particularly in the adolescent and elderly population. Clinically, there are also other non-bony visual observations on CXRs, such as various medical implants and devices, that occur more commonly in older people compared to younger people. In general, radiologists also seem to have a rough visual estimate about a patient's age when reading CXRs. In practice, as the most commonly requested modality in a healthcare encounter, giving each patient feedback on his or her computer estimated CXR age could possibly be an interesting public health intervention aimed at prompting people to adopt healthier behaviors.

In this work we report a solution for automatic estimation of patient age from CXR images. We achieve this by training a deep neural network on a large public dataset. Since CXR images have not been widely used for age estimation in the computer vision literature, we do not already have established imaging features known to be good indicators of age, though we do have good biological reasons to expect the presence of these features. Therefore, deep learning, which provides a platform for learning features, is a good solution here. Convolutional Neural Networks (CNNs) have witnessed an explosion in use cases over the past 10 years with unprecedented accuracies in various domains including medical image processing. Their success lies in their ability to learn a plethora of features in the most optimal way for each specific problem. As a downside, they require large amounts of labeled data to train and optimize on. However, the recent availability of large clinical CXR datasets have enabled us to use CNNs for this age estimation work [8].

While our methodology uses the whole image as the input, understanding what parts of the image contribute to age determination can be insightful. For this, we propose to use activation maps. Activation mapping is a technique in deep learning to visualize the attention of a CNN with respect to output classes by generating heat maps. It helps us understand which local areas of an input image drive the network to decide. This approach has been extensively used in numerous applications to check whether the network learns the proper parts of an image to infer its decision. Our work differs from previous works in this area in 3 key points: 1) We utilized a large public dataset to predict a patient's overall biological age incorporating multiple visualized anatomical structures, 2) we used CXRs versus previous established modalities, and finally 3) we utilized activation maps to get an insight into which areas of the CXR have guided our deep network to predict patient age. To our knowledge this is the first work that attempts to predict a patient's biological age from CXRs through a computer algorithm.

## 2. METHODS AND MATERIALS

### 2.1 Dataset

For this work, we used a popular publicly available dataset: the NIH ChestX-ray8 [8]. It contains more than 110,000 frontal CXR images from 30,000 unique individuals. It also comes with metadata containing the following information for each image: 1) Image Index, 2) Finding Labels, 3) Follow-up Visit Number, 4) Patient ID, 5) Patient Age, 6) Patient Gender, 7) View Position, 8) Original Image Width and Height, 9) Original Image Pixel Spacing. For our work, we were interested in predicting the patient's age, and we further explored the age prediction performance for each view position.

## 2.2 Network architecture and training

For the CNN architecture we used DenseNet, which has proven to be very successful in various computer vision classification problems including CXR imaging applications [9], [10], [11] and [12]. DenseNet has several compelling advantages: they alleviate the vanishing-gradient problem, strengthen feature propagation, encourage feature reuse, and substantially reduce the number of parameters [13].

We used DenseNet-169 as our network architecture was pre-trained on ImageNet. The choice of this DenseNet version was intentional because we have witnessed it performing very well in comparable tasks (e.g. CXR finding classification). However, when using other popular DenseNet architectures (i.e. 121 and 201) we didn't notice any significant change in our age assessment results. As the original architecture is a classifier, we reformulated age prediction as a regression problem. This involved modifying the network's architecture by replacing the last connected layer with a connected layer that had a single-neuron (i.e. 1 output neuron). The activation for this node was set to 'Sigmoid' (i.e. range of 0 to 1).

Figure 1 shows the raw distribution of patient age for the whole dataset after removing just 19 outliers that had values above 90 (years old). In order to perform regression, we subsequently normalized the patient age to within the range of 0 to 1 by dividing all age values by the maximum age value (i.e. 90).

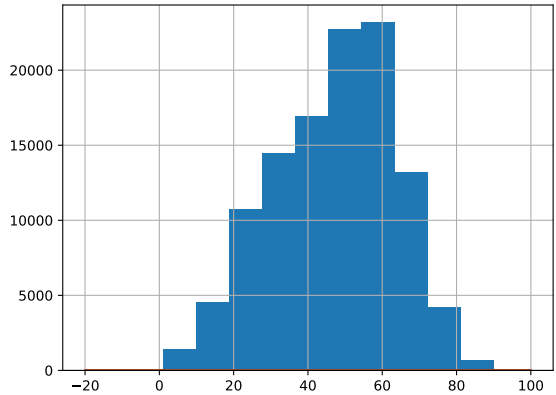

Figure 1. Age distribution for NIH ChestX-ray8 dataset

We split the dataset into 80% training and 20% validation. No data augmentation was applied. We trained the network until validation loss plateaued. We also calculated the coefficient of determination, $R^2$, because in dealing with a regression problem we wanted to identify the goodness-of-fit of the trained network. $R^2$ is a statistical measure of how close the data are to the fitted regression line [15] and it tends to be a good indicator for regression model fitting. It is calculating by the following formula: $R^2 = 1 - \frac{SSresidual}{SStotal}$ where **SSresidual** is the sum of squares of residuals and **SStotal** is the total sum of squares.

After network training completion, we used activation maps to examine where the network is activated for the age prediction task. To generate these activation maps we used "keras-vis" package [16] because it offered a variety of activation mapping methodologies, such as activation maximization and saliency maps. We used the saliency method of Simonyan et. al. described in [17] to generate these activation maps.

# 3. EXPERIMENTS AND RESULTS

## 3.1 CXR Age estimation results

As we mentioned in the previous paragraph, we used DenseNet-169 and we performed 3 experiments for each CXR view position: Anterior Posterior (AP, patient is in anteroposterior position), Posterior Anterior (PA, patient is in posterioranterior position) and combined frontal views (PA+AP). This is because we know that CXR view affects organ appearance due to minor variation of X-ray beam projection. In addition, the underlying patient populations likely have different distributions of health status or diseases because AP CXRs are more often requested for critically ill patients, who often cannot physically tolerate a PA CXR exam, which are acquired from an upright standing position. Therefore, we also wanted to explore how this view position factor affects age prediction. Figure 2 below shows a random AP and PA CXR for comparison of disease burden.

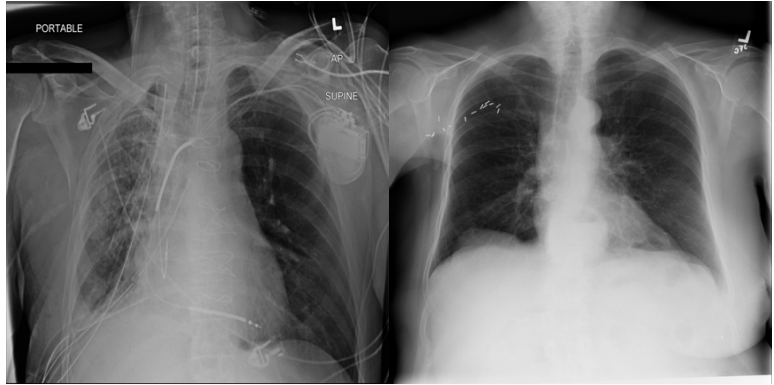

**Figure 2. AP view of patient with ID 00000013 (left) and PA view of patient with ID 00000003 (right)**

Figure 3 shows the model's loss for the 3 experiments.

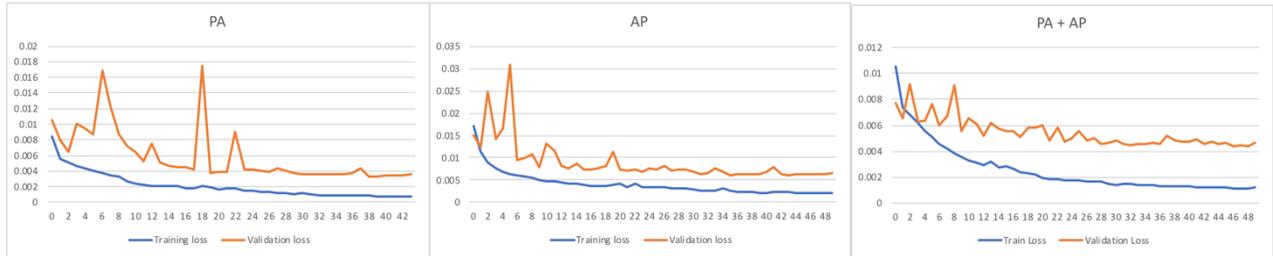

**Figure 3. Training and validation loss for each type of experiment (from left to right: PA, AP and PA+AP). Minimum validation loss: PA: 0.003, AP: 0.006, PA+AP: 0.004. We observed some overfit for the PA+AP experiment**

Below, figure 4 shows the $R^2$ value (i.e. coefficient of determination) for each experiment as well.

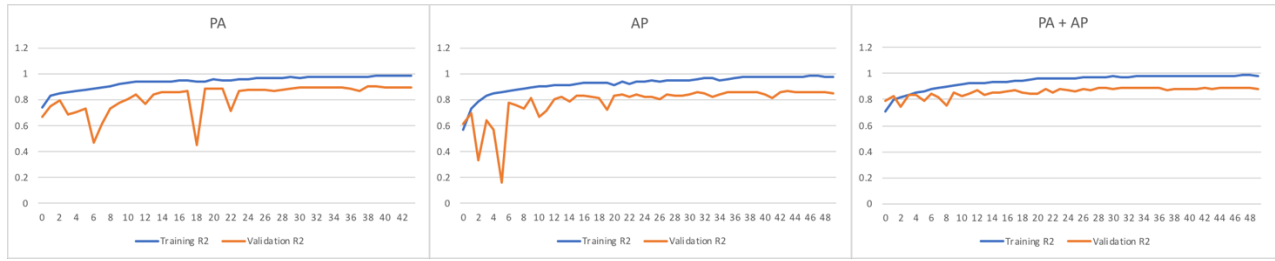

**Figure 4.** Training and validation $R^2$ value for each type of experiment (from left to right: PA, AP and PA+AP). Best validation $R^2$: PA: 0.90, AP: 0.86, PA+AP: 0.89

**Table 1 Recall (Percentage of cases predicted correctly within the age error ranges)**

|  | PA | AP | PA+AP |
|---|---|---|---|
| ± 4 years | **0.6745** | **0.5896** | **0.6469** |
| ± 9 years | **0.9441** | **0.8856** | **0.9277** |

Table 1 shows the recall for each view position and for two different age error ranges centered on the real age: -4 to +4 years and -9 to +9 years. We get the best recall for the PA view using the ± 9 years error range. Our hypothesis that age prediction is more difficult in AP view is verified by its lower score, which also pushes the combined PA+AP model's recall down.

We made two other observations during training: 1) Our network tends to slightly over estimate the patient's age. 2) Our attempt at training the networks with the $R^2$ loss (i.e. coefficient of determination) as an objective function instead of L2 did not show any major improvement in recall or $R^2$.

### 3.2 Age heat map analysis

We visualized the network to understand how it produces its regression output (i.e. age: 0-1). For this we used the saliency map as we mentioned in paragraph 2.2. This, in return, tells us which parts of the input image contributed towards a change of output (i.e. age) [17]. In the figures below, examples of the network's activation are shown for a few CXRs with age correctly predicted within ± 9 years.

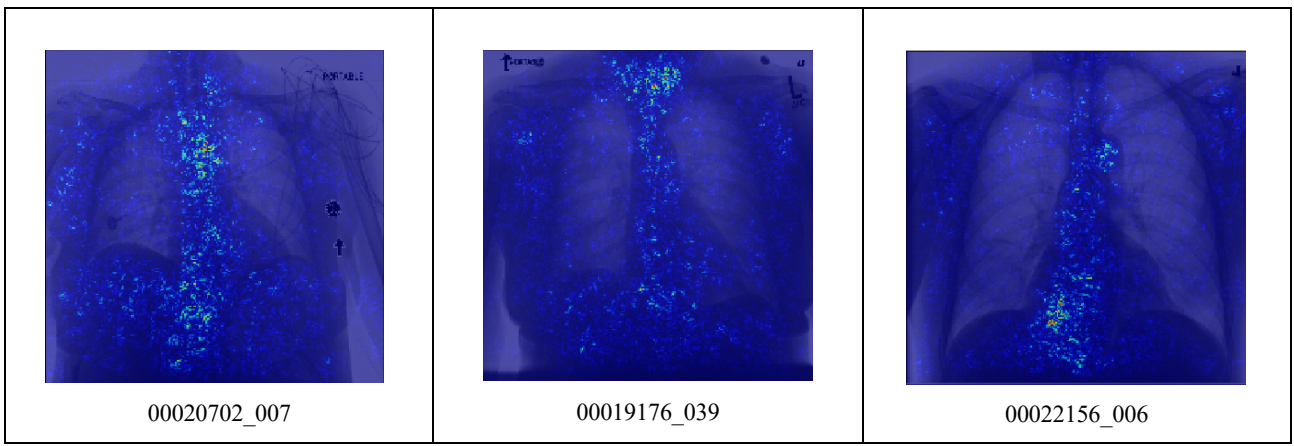

| 00020702_007 | 00019176_039 | 00022156_006 |

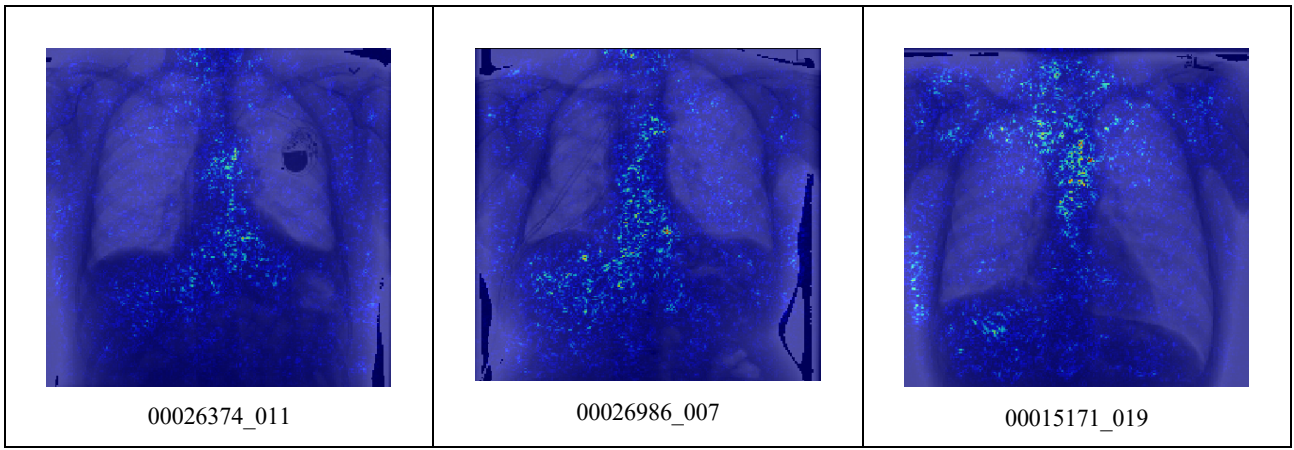

Figure 5. Examples for patients that are older than 65 years (original image name for reference)

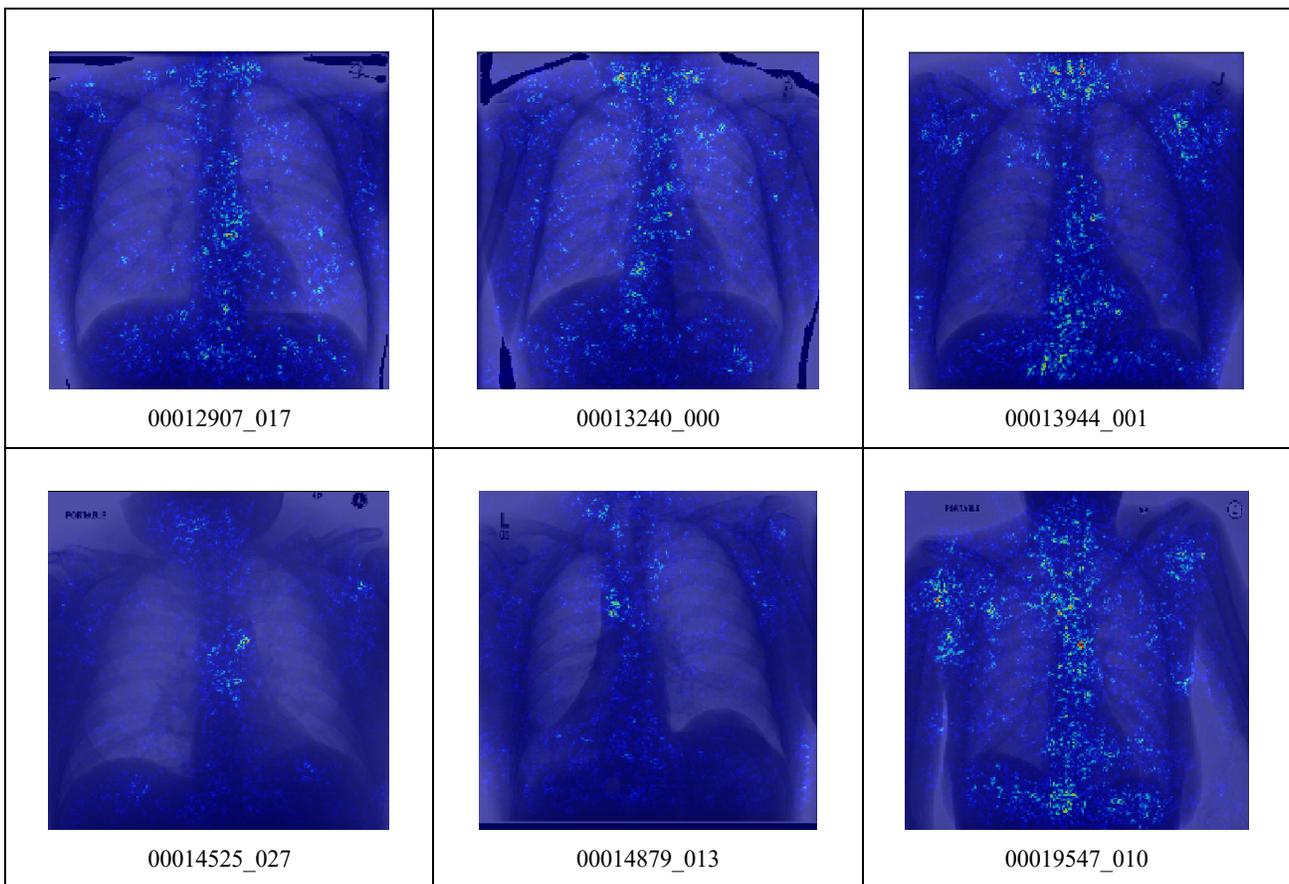

Figure 6. Activation maps for patients between 25 year old and 65 years old (original image name for reference)

From figures 5 and 6 above we noticed some interesting results: The network highlights mostly the following areas: neck, clavicles, mediastinum, ascending aortic arch and spine, where we would expect age related change in appearance biologically. We also noticed greater heterogeneity in regional activation for younger patients (25 < real age <65 years) versus older patients (real age > 65 years). That is, the lungs and the bony and joint regions (clavicles, shoulders and spine) were more likely to be salient for younger patients.

### 3.3 Age prediction disparity analysis

We also examined extreme cases of disparity between the real and predicted age. More specifically, these cases were the ones that the network's predicted age values that deviated more than 10 years from the real age of the patient. The reason for this analysis was because we wanted to visually examine what anatomical or pathological features might have led the network to predict a very different age compared to a patient's real age. Table 2 and 3 below show some cases and qualitative comments from a radiologist.

Table 2. Three examples the algorithm predicted a patient's age much smaller than real age

| Sample images predicted younger (image name) | Real Age | Predicted Age | Radiologist's Guess | Reason given by radiologist |
|---|---|---|---|---|
| 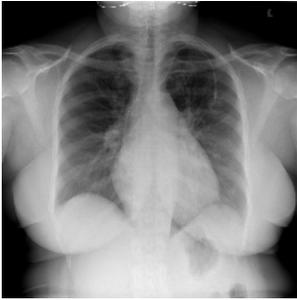 00022793_000.png | 70 | 50 | 30-50 | Fairly clear lungs without interstitial prominence. The spine and shoulder joints don't look degenerated. Normal heart size. |
| 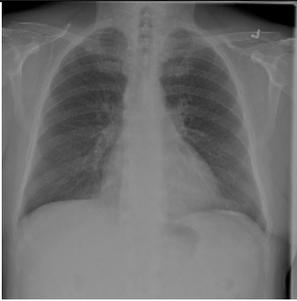 00001824_000.png | 76 | 61 | 60-70 | Some more interstitial prominence than a middle age person. Only has mild spinal degeneration visible. |
| 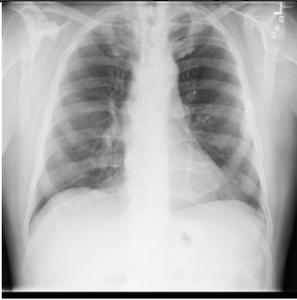 00017483_000.png | 78 | 64 | 50-70 | The picture quality on this one makes the possible age range wider. But the lungs are fairly clear. Possible old left rib fracture or pleural based mass, but bones otherwise look healthy. Some age-related spinal degeneration and possible aortic calcifications visible. |

Table 3. Three examples the algorithm predicted a patient's age much larger than real age

| Sample images predicted older | Real Age | Predicted Age | Radiologist's Guess | Reason given by radiologist |
|---|---|---|---|---|
| 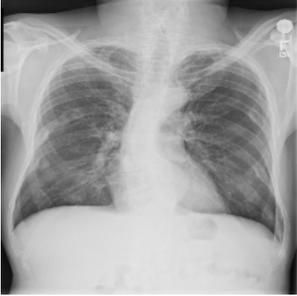<br>00003421_001.png | 41 | 56 | 70+ | Tortuous aorta, out of proportion to the mild scoliosis, as well as suboptimal body positioning, both of which often indicate an older and less mobile patient. There's interstitial prominence, right greater than left. |
| 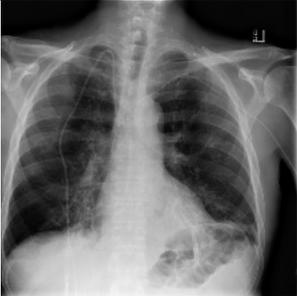<br>00017236_028.png | 46 | 60 | 60+ | This level of severity of lung lucency consistent with chronic obstructive lung disease, suggest the patient is likely at least 60. There's also pleural scarring at the left lung base, which is more common in older patients. |
| 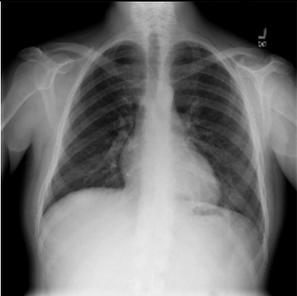<br>00010434_000.png | 21 | 35 | 30-50 | Hard to explain why. All the anatomy appears normal, but patient appears older than 30; a younger patient would just typically look "cleaner" than this image. |

Overall, our observations of images with a large real-predicted age gap seem to suggest that the disparities may be related to either image quality, or the presence or lack of age-related diseases or changes that are expected for a patient's real age group.

## CONCLUSION

In this paper we explored age estimation of patients using frontal CXRs. For this we used a large public dataset and a deep convolutional network. The network was trained in regression fashion. We managed to achieve accuracies up to 94% for PA CXRs (*i.e.* posterior to anterior view position). We used activation maps to highlight the areas of the image that contributed to the age estimation, which has not been done before.

This work set the stage for the next research directions we would like to follow. More specifically, although the activation maps can help visualize areas predictive of age in CXRs, the real clinical value is to explore how we can utilize these maps in disease classifiers so that we can identify how much of an abnormal region is due to the natural age progression and how much is due a pathology/disease. We believe this extra information can help build better disease classifiers, such as classifying whether an image is normal or abnormal "for age". The information may also help quantify the degree of physiologic and pathologic processes present in patients. Examples of the latter may include but are not limited to osteoarthritis or degenerative disc disease of the musculoskeletal system, senescent or emphysematous changes of the lungs, and calcific atherosclerosis of the aorta and vasculature. Indeed, as algorithms improve it is possible that previously unknown but more reliable identifiers of aging or pathology will be identified. In the clinical setting, a physician may find in the predicted age of various organ systems or the overall predicted age, the opportunity to counsel the patient with respect to improving health habits.